\documentclass[conference]{IEEEtran}
\IEEEoverridecommandlockouts
\usepackage{cite}
\usepackage{amsmath,amssymb,amsfonts}
\usepackage{algorithmic}
\usepackage{graphicx}
\usepackage{textcomp}
\usepackage{xcolor}
\usepackage{braket}
\usepackage{authblk}
\def\BibTeX{{\rm B\kern-.05em{\sc i\kern-.025em b}\kern-.08em
    T\kern-.1667em\lower.7ex\hbox{E}\kern-.125emX}}
\begin{document}

\title{Quantum Pointwise Convolution: A Flexible and Scalable Approach for Neural Network Enhancement}

\author[1]{An Ning}
\author[2]{Tai Yue Li}
\author[3]{Nan Yow Chen}

\affil[1]{School of Electrical Engineering, KAIST, Daejeon, South Korea}
\affil[2]{National Synchrotron Radiation Research Center (NSRRC), Hsinchu, Taiwan}
\affil[3]{National Center for High-Performance Computing (NCHC), Hsinchu, Taiwan}

\maketitle

\begin{abstract}

In this study, we propose a novel architecture, the Quantum Pointwise Convolution, which incorporates pointwise convolution within a quantum neural network framework. Our approach leverages the strengths of pointwise convolution to efficiently integrate information across feature channels while adjusting channel outputs. By using quantum circuits, we map data to a higher-dimensional space, capturing more complex feature relationships. To address the current limitations of quantum machine learning in the Noisy Intermediate-Scale Quantum (NISQ) era, we implement several design optimizations. These include amplitude encoding for data embedding, allowing more information to be processed with fewer qubits, and a weight-sharing mechanism that accelerates quantum pointwise convolution operations, reducing the need to retrain for each input pixels. In our experiments, we applied the quantum pointwise convolution layer to classification tasks on the FashionMNIST and CIFAR10 datasets, where our model demonstrated competitive performance compared to its classical counterpart. Furthermore, these optimizations not only improve the efficiency of the quantum pointwise convolutional layer but also make it more readily deployable in various CNN-based or deep learning models, broadening its potential applications across different architectures.

\end{abstract}


\section{Introduction}
Quantum machine learning is an emerging field that combines quantum algorithms with classical machine learning models to enhance performance. Quantum algorithms exploit phenomena such as superposition and entanglement to process information in ways inaccessible to classical algorithms. Various approaches have been proposed in this domain. For example, Quantum Support Vector Machines (QSVM) \cite{b1}-\cite{b2} leverage quantum algorithms to optimize support vector machines, improving efficiency in high-dimensional classification tasks. Quantum Convolutional Neural Networks (QCNN) \cite{b3}-\cite{b5}, inspired by classical CNNs, implement convolutional, pooling, and fully connected layers using quantum circuits. Furthermore, parameterized quantum circuits have been applied to develop quantum machine learning models \cite{b6}-\cite{b8}, with applications such as high-energy physics event classification demonstrating quantum advantages \cite{b9}-\cite{b10}.

In classical convolutional neural networks, pointwise convolution, or 1×1 convolution, employs a 1×1 kernel to exclusively process channel-wise relationships at individual spatial positions, thereby modifying feature channel interactions without affecting spatial dimensions. It is a key component in models like MobileNet \cite{b11}, where it combines with depthwise convolution to create depthwise separable convolutions. This two-step method significantly reduces computational complexity while preserving performance, making it suitable for resource-constrained environments such as mobile and embedded systems. Although pointwise convolution is effective for channel manipulation, its linear nature limits its capacity to capture complex nonlinear features.

To address this limitation, nonlinear activation functions \cite{b12}-\cite{b15} are typically applied directly after the pointwise convolution layer. This combination enhances the expressive power of the model, enabling it to learn and capture more complex and high-dimensional feature representations. However, even with activation functions, classical pointwise convolution remains constrained by the inherent limitations of classical computation in modeling complex nonlinear relationships.

Motivated by these challenges and inspired by advancements in quantum machine learning, we propose utilizing quantum circuits to implement pointwise convolution, thereby overcoming these constraints. Quantum neural networks employ various encoding techniques to efficiently map classical data into high-dimensional quantum Hilbert spaces, leveraging the unique properties of quantum systems. By incorporating parameterized quantum gates as quantum convolution operations, the network performs nonlinear transformations on input data. Furthermore, quantum entanglement allows for enhanced interactions between channels, capturing intricate feature relationships that classical methods struggle to model.

This novel approach retains the advantages of classical pointwise convolution while significantly enhancing its capacity to model complex feature relationships. By integrating quantum circuits into the pointwise convolution operation, we aim to bridge the gap between classical and quantum computation, addressing the limitations of traditional methods and boosting overall network performance.

In this study, we employ the FashionMNIST and CIFAR10 datasets to evaluate convolutional neural networks incorporating both quantum and classical pointwise convolution for multi-label classification. This framework enables a direct performance comparison between quantum and classical pointwise convolution. For the quantum model, amplitude encoding is utilized to input data into the quantum circuit, ensuring seamless integration with the classical model. The quantum model is implemented using the noiseless simulator provided by Pennylane \cite{b17}, while the classical model is constructed with PyTorch.

This article is structured as follows: Section 2 outlines the architecture and principles of classical pointwise convolution, along with the foundational concepts of quantum machine learning. Section 3 describes the integration of quantum computing with pointwise convolution, covering data encoding, circuit design, execution, measurement, and the generation of output feature maps. Section 4 presents experimental results on multiple datasets, followed by a discussion of potential optimizations and applications.

\section{Prelliminaries}

\subsection{Pointwise Convolution}
Pointwise convolution \cite{b11} is a specialized operation in convolutional neural networks (CNNs) that processes each spatial position across multiple channels in the input tensor while preserving spatial resolution. Unlike traditional convolutions, which operate over local spatial neighborhoods of the input, pointwise convolution focuses solely on the channel-wise interaction at each pixel. It is often used to modify the number of channels, which makes the computation more efficient.

Given an input tensor $\mathbf{X} \in \mathbb{R}^{H \times W \times C_{\text{in}}}$, where $H$ and $W$ represent the spatial dimensions (height and width), and $C_{\text{in}}$ denotes the number of input channels, a pointwise convolution is applied using a kernel $\mathbf{W} \in \mathbb{R}^{1 \times 1 \times C_{\text{in}} \times C_{\text{out}}}$, where $C_{\text{out}}$ is the number of output channels. The $1 \times 1$ convolution kernel indicates that the operation only affects the channel dimension for each spatial location $(i, j)$, without considering adjacent pixels. The output tensor $\mathbf{Y} \in \mathbb{R}^{H \times W \times C_{\text{out}}}$ is computed as:

\[
Y_{i,j,k} = \sum_{c=1}^{C_{\text{in}}} X_{i,j,c} \cdot W_{c,k}
\]
where $i$ and $j$ represent the spatial dimensions, $c$ is the input channel index, and $k$ is the output channel index. This formula shows that for each spatial location $(i, j)$, the value of each output channel is the weighted sum of the corresponding input channel values. This operation allows the network to compress or expand information across the channel dimension while maintaining the spatial structure of the input (i.e., the height and width $H \times W$ remain unchanged).

\subsection{Framework for Quantum Machine Learning}

\subsubsection{Data Embedding, i.e., Classical to Quantum}
In the quantum pointwise convolution layer, classical data is embedded into quantum circuits using amplitude encoding. Amplitude encoding \cite{b16} is a general method for embedding classical data into quantum states, associating classical data with the probability amplitudes of a quantum state. Given an input vector $\mathbf{x} \in \mathbb{R}^{N}$, amplitude encoding maps it to the quantum state of $n$ qubits $\ket{\phi(x)}$, which is expressed as:

\[
U_{\phi}(x): x \in \mathbb{R}^N \rightarrow \ket{\phi(x)} = \frac{1}{\|\mathbf{x}\|} \sum_{i=1}^{N} x_i \ket{i}
\]
where $\ket{i}$ represents the $i$-th computational basis, and $\|\mathbf{x}\|$ is the norm of the input vector. The advantage of this encoding method is that quantum computing can represent exponentially large classical data, significantly improving algorithm efficiency in practice.

\subsubsection{Quantum Circuit and execution}
In quantum machine learning, designing the quantum circuit is critical, involving the selection of quantum gates to manipulate qubits for the desired functionality. The circuit starts with quantum state preparation, where qubits are initialized to represent input data using methods like amplitude or basis encoding. Quantum gates are then applied sequentially to perform transformations and interactions, followed by measurement to extract classical results for training.

\subsubsection{Parameter Updates, Loss Function, and Training}
Training in quantum machine learning involves defining a loss function to measure the difference between model predictions and actual outcomes, with common choices like cross-entropy or Mean-Square-Error (MSE). Classical optimization algorithms, such as gradient descent, adjust the circuit parameters using gradient estimation techniques like the parameter-shift rule or finite differences. The training loop iteratively updates parameters to minimize the loss and improve performance, with measurement outcomes post-processed for meaningful interpretation.

\begin{figure*}[t]
    \centering
    \includegraphics[width=\textwidth]{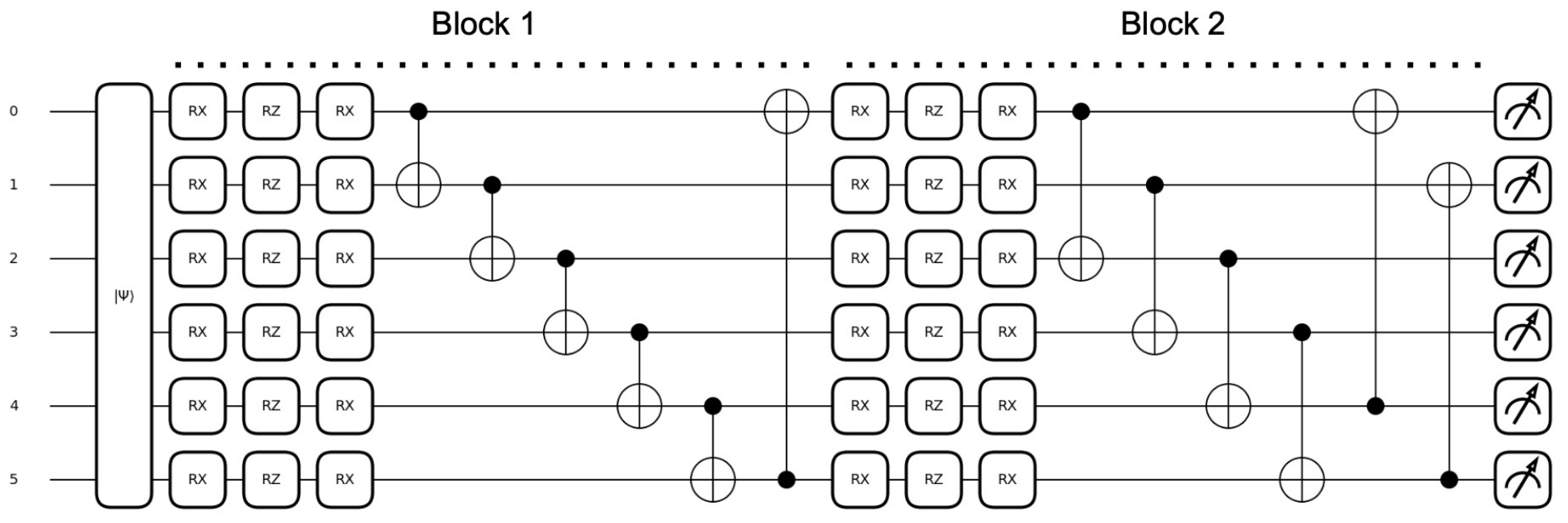}  
    \caption{The structure of quantum circuit with strongly entanglement layers (Blocks). The circuit first employs amplitude encoding, where the classical input data is embedded into the quantum state by varying the amplitude of the quantum states across the 6 qubits. Then, apply the strongly entangling circuit architecture, designed with 6 qubits and organized into 2 sequential blocks $B_1$ and $B_2$. Each block consists of single-qubit gates $R_X$ and $R_Z$ rotations and CNOT gate applied to every qubit, ensuring that each qubit undergoes individual quantum operations. Each qubit’s state is measured using Pauli-Z operators}
\end{figure*}

\section{ Quantum Pointwise Convolution} 
In this section, we introduce quantum pointwise convolution, which is completed by the following steps.

\subsection{Data Preparation and Embedding}
For each pixel, the data across all channels is aggregated into a vector and normalized before being embedded into a quantum state using amplitude encoding. This process ensures that the input data is mapped to a valid quantum state, represented as a vector \( |\psi\rangle \) in the Hilbert space of the quantum system.

During normalization, the vector is rescaled to satisfy the quantum state's requirement of being a unit vector. Specifically, the normalization ensures that the vector satisfies \( \langle \psi | \psi \rangle = 1 \), a fundamental property of quantum states.

\subsection{Quantum Circuit Architecture}

In this study, we use strongly entangling circuits inspired by circuit-centric classifier design \cite{b18} for quantum pointwise convolution. These circuits project an encoded feature vector \(\psi\) into the Hilbert space of an \(n\)-qubit quantum system, transforming it into a quantum state. The quantum circuit then applies a parameterized unitary operator \(U(\theta)\), defined by a set of variables \(\theta\), to convert \(\psi\) into another quantum state \(\psi' = U(\theta) \psi\). By generating entangled states across multiple qubits, the circuits enable complex transformations in a high-dimensional Hilbert space.

To realize the unitary operator \(U(\theta)\), we decompose it as a sequence of single-qubit and two-qubit gates:

\[
U = U_L \dots U_\ell \dots U_1,
\]
where each \(U_\ell\) represents either a single-qubit or a two-qubit gate. For a single-qubit gate \(G_k\) acting on the \(k\)-th qubit in a system of \(n\) qubits, it can be expressed as:

\[
U_\ell = I_0 \otimes \dots \otimes G_k \otimes \dots \otimes I_{n-1},
\]
where \(I_i\) denotes the identity operation on the \(i\)-th qubit. This decomposition allows us to implement flexible unitary transformations, projecting input data into a high-dimensional quantum feature space.

\subsubsection{Single-Qubit Gates}

In the quantum circuit, each single-qubit gate is typically composed of a sequence of rotations, represented as:

\[
G_j = R_z(\mu_1) R_x(\mu_2) R_z(\mu_3),
\]
where \(\mu_1\), \(\mu_2\), and \(\mu_3\) are parameterized angles that are optimized during training. The rotation matrices \(R_x(\theta)\) and \(R_z(\theta)\) are defined as follows:

\[
R_x(\theta) = \begin{pmatrix} \cos\left(\frac{\theta}{2}\right) & -i\sin\left(\frac{\theta}{2}\right) \\ -i\sin\left(\frac{\theta}{2}\right) & \cos\left(\frac{\theta}{2}\right) \end{pmatrix},
\]
\[
R_z(\theta) = \begin{pmatrix} e^{-i\frac{\theta}{2}} & 0 \\ 0 & e^{i\frac{\theta}{2}} \end{pmatrix}.
\]

These rotation gates allow controlled evolution of the quantum state along the \(X\) and \(Z\) axes of the Bloch sphere.

\subsubsection{Two-Qubit Gates}

To introduce quantum entanglement, we utilize two-qubit gates to create non-classical correlations between qubits. We employ two-qubit gates that can transform two-qubit product states into entangled states, with the CNOT (Controlled-NOT) gate being a common example. For a two-qubit gate \(C(G)\), acting between control qubit \(a\) and target qubit \(b\), we define:

\[
C_a(G_b) \ket{x}\ket{y} = \ket{x} \otimes G^{x}_b \ket{y},
\]
where \(G\) is a single-qubit gate applied to the target qubit \(b\), conditioned on the state \(x\) of the control qubit \(a\). 

By alternating between single-qubit and two-qubit gates, the circuit is able to establish entanglement across qubits, which is critical for modeling complex correlations within the data.

\subsubsection{Full Block Composition}

The complete circuit layer comprises a series of single-qubit rotation gates followed by two-qubit entangling gates. A single block, denoted as \(B\), can be expressed as following, as shown in \cite{b18}:

\[
B = \prod_{k=0}^{n-1} R^X_k C_{ck}(P_{tk}) \prod_{j=0}^{n-1} G_j,
\]
where \(R^X_k\) is an \(X\)-axis rotation on qubit \(k\), \(C_{ck}(P_{tk})\) is a controlled phase gate with a phase shift \(P_{tk}\), and 
\(G_j\) is a single-qubit gate represented by the sequence \(R_z(\mu_1) R_x(\mu_2) R_z(\mu_3)\).

In Fig.1, we show the example of structure of quantum circuit with 6-qubits and two strongly entanglement layers (blocks).

This structure allows the quantum circuit to process input data effectively by creating both short-range and long-range entanglement, enabling the system to capture complex patterns that may be difficult for classical networks to model.

\subsection{Quantum Circuit Processing}
After embedding classical data into a quantum state, the quantum circuit processes the data through layers of strongly entangling circuits. Each layer applies single-qubit rotations and multi-qubit entangling gates, transforming the quantum state to capture complex, expressive features based on the input data.

The overall unitary operation of the quantum circuit can be expressed as:
\[
U(\theta) = \prod_{l=1}^L B_l(\theta_l),
\]
where $L$ is the number of layers (code blocks), $B_l$ is the unitary operation of the $l$-th code block, $\boldsymbol{\theta}_l$ are the parameters (rotation angles) for the $l$-th code block.

The parameters of the quantum circuit (the rotation angles of the gates) are trained using gradient-based optimization techniques "parameter-shift rule"\cite{b19}. The gradient of an expectation value $\langle O \rangle$ with respect to a parameter $\theta$ is given by:
\[
\frac{\partial \langle O \rangle}{\partial \theta} = \frac{1}{2} \left( \langle O \rangle_{\theta + \frac{\pi}{2}} - \langle O \rangle_{\theta - \frac{\pi}{2}} \right).
\]
This approach allows for the efficient calculation of gradients in quantum circuits, facilitating the integration of the quantum circuit into an end-to-end trainable model. By optimizing these parameters, the quantum circuit learns to perform transformations that enhance feature extraction beyond the capabilities of classical methods.

\subsection{Measurement}
After the quantum circuit processes the input data, each qubit's state is measured using Pauli-$Z$ operators. The measurement yields the expectation value $\langle Z_i \rangle$ for each qubit, where $Z_i$ represents the Pauli-$Z$ operator acting on the $i$-th qubit. The expectation value for each qubit is calculated as:
\[
\langle Z_i \rangle = \langle \psi_{\text{out}} | Z_i | \psi_{\text{out}} \rangle,
\]
where $|\psi_{\text{out}}\rangle = U(\theta) |\psi \rangle$ is the final state of the quantum system after processing through the circuit.

\subsection{Multiple Quantum Pointwise Convolutional Kernels}

In quantum pointwise convolution, each quantum circuit processes input pixels across multiple channels and generates new feature maps. The number of feature maps produced by a quantum circuit corresponds to the number of qubits utilized. To replicate the effect of multiple convolutional kernels, multiple quantum circuits are employed, with each circuit functioning as a distinct convolutional kernel, analogous to classical convolutional neural networks.

\subsubsection{Feature Map}

Given an input tensor $\mathbf{X} \in \mathbb{R}^{H \times W \times C_{\text{in}}}$, where $H$ and $W$ are the height and width of the image and $C_{\text{in}}$ is the number of input channels (features), quantum pointwise convolution processes each pixel across channels. Each quantum circuit outputs feature maps that correspond to the number of qubits, $n_{\text{qubits}}$. For example, a quantum circuit with $n_{\text{qubits}} = 6$ can generate 6 output feature maps. 

For each pixel at position $(i, j)$ in the input image, we can represent the output feature map from a single quantum circuit as:

\[
\mathbf{F}_{i,j} = f_{\text{quantum}}\left( \mathbf{X}_{i,j,:} \right)
\]
where $\mathbf{F}_{i,j} \in \mathbb{R}^{n_{\text{qubits}}}$ represents the output feature vector for pixel $(i, j)$, and $f_{\text{quantum}}$ represents the quantum circuit that processes the input pixel’s channel data $\mathbf{X}_{i,j,:} \in \mathbb{R}^{C_{\text{in}}}$.

\begin{figure*}[!htbp]
    \centering
    \includegraphics[width=\textwidth]{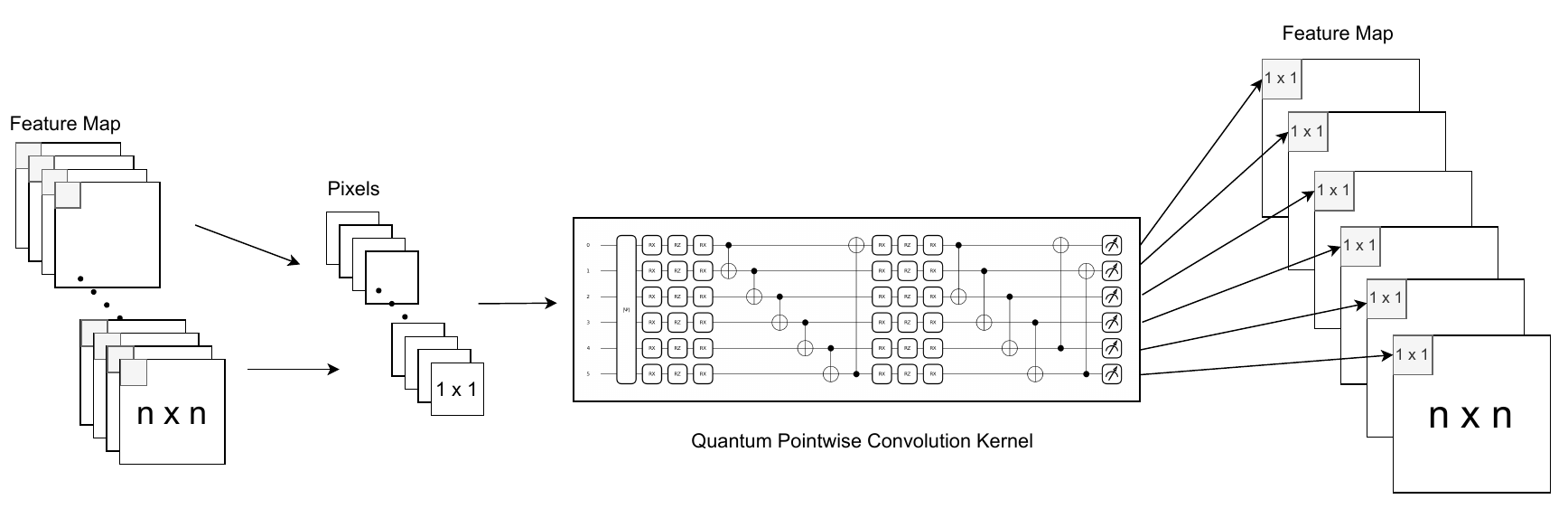}  
    \caption{The structure of a quantum pointwise convolutional layer. From a multi-channel feature map, collect pixel values at the same position across channels and concatenate them into a vector. Feed this vector into quantum pointwise convolution kernels. After measuring each qubit, assign the measurement values back to the corresponding position in the new feature map. By iterating through the entire feature map, multiple new feature maps will be generated, depending on the number of qubits and kernels used.}
\end{figure*}

Each quantum circuit outputs multiple feature maps based on the number of qubits. The total number of output feature maps from the quantum layer is determined by the number of quantum circuits employed. If there are $n_{\text{circuits}}$ quantum circuits, the total number of output feature maps is:

\[
C_{\text{out}} = n_{\text{circuits}} \times n_{\text{qubits}}
\]

Thus, using multiple quantum circuits simulates the concept of having multiple convolutional kernels, with each circuit producing a set of feature maps corresponding to its qubits.

\subsubsection{Concatenation}

Once all the quantum circuits have processed the input tensor, their outputs are concatenated to form the final output tensor. Let $\mathbf{F}^{(k)}$ represent the feature maps produced by the $k$-th quantum circuit, where $k \in \{1, 2, \dots, n_{\text{circuits}}\}$. The final output tensor is obtained by concatenating all the feature maps along the channel dimension:

\[
\mathbf{Y}_{i,j} = \left[\mathbf{F}_{i,j}^{(1)}, \mathbf{F}_{i,j}^{(2)}, \dots, \mathbf{F}_{i,j}^{(n_{\text{circuits}})} \right]
\]
where $\mathbf{Y}_{i,j} \in \mathbb{R}^{C_{\text{out}}}$ is the output feature vector for pixel $(i, j)$, and $C_{\text{out}} = n_{\text{circuits}} \times n_{\text{qubits}}$.

\subsubsection{Multiple Kernels}

In classical CNNs, convolutional layers apply multiple kernels (filters) to an input tensor to extract features, with each kernel generating a feature map. The aggregated feature maps form the output tensor. In quantum pointwise convolution, this process is replicated using multiple quantum circuits.

Each quantum circuit with \( n_{\text{qubits}} \) serves as a convolutional kernel, producing \( n_{\text{qubits}} \) feature maps:
\begin{itemize}
    \item The use of \( n_{\text{circuits}} \) quantum circuits mimics multiple convolutional kernels, resulting in \( n_{\text{circuits}} \times n_{\text{qubits}} \) total feature maps.
    \item Quantum pointwise convolution incorporates \textbf{weight sharing}, where the parameters (rotation angles) of each quantum circuit, analogous to kernel weights in classical convolution, are shared across the input. This reduces the parameter count while enabling the capture of spatial patterns across different input regions.
\end{itemize}

Quantum pointwise convolution thus emulates classical convolutional layers by using quantum circuits as kernels to generate feature maps, offering enhanced efficiency and potential advantages over classical pointwise convolution.

In Figure 2, the process is illustrated where each quantum circuit, with multiple layers and qubits, produces feature maps for a given set of input pixels, analogous to classical convolutional kernels, with the added benefit of weight sharing across the input.

\subsection{Loss Function}

In training the quantum pointwise convolutional neural network, the cross-entropy loss function is employed to quantify the discrepancy between predicted outputs and true labels. This loss function is particularly suitable for classification tasks, as it evaluates the divergence between the predicted probability distribution and the actual distribution represented by the labels.

Given a dataset with \( N \) samples, where each sample has an input \( \mathbf{x}^{(i)} \) and a corresponding true label \( \mathbf{y}^{(i)} \), the cross-entropy loss \( \mathcal{L} \) is defined as:

\[
\mathcal{L} = -\frac{1}{N} \sum_{i=1}^{N} \sum_{k=1}^{K} y^{(i)}_k \log \hat{y}^{(i)}_k,
\]
where \( K \) is the number of classes, \( \mathbf{y}^{(i)} = [y^{(i)}_1, y^{(i)}_2, \dots, y^{(i)}_K] \) is the one-hot encoded true label vector for the \( i \)-th sample, \( \hat{\mathbf{y}}^{(i)} = [\hat{y}^{(i)}_1, \hat{y}^{(i)}_2, \dots, \hat{y}^{(i)}_K] \) is the predicted probability vector for the \( i \)-th sample, obtained from the model's output.

The predicted probabilities are computed by applying the softmax function to the logits \( \mathbf{z}^{(i)} \) produced by the model:

\[
\hat{y}^{(i)}_k = \frac{\exp(z^{(i)}_k)}{\sum_{j=1}^{K} \exp(z^{(i)}_j)},
\]
where \( z^{(i)}_k \) is the logit (unnormalized output) for class \( k \) of the \( i \)-th sample.

In the context of the quantum pointwise convolutional neural network, the logits \( z^{(i)} \) are derived from the measurements of the quantum circuit. After processing the input data through the quantum circuits and obtaining the expectation values \( \langle Z_q \rangle \) for each qubit \( q \), these values are possibly passed through additional classical layers to produce the final logits:

\[
\mathbf{z}^{(i)} = f_{\text{classical}}(\langle Z \rangle^{(i)}),
\]
where \( \langle Z \rangle^{(i)} = [\langle Z_1 \rangle^{(i)}, \langle Z_2 \rangle^{(i)}, \dots, \langle Z_{n_{\text{qubits}}} \rangle^{(i)}] \) is the vector of expectation values for the \( i \)-th sample, and \( f_{\text{classical}} \) represents any additional classical computation applied to the quantum outputs.

By minimizing the cross-entropy loss function, we optimize the parameters of the quantum circuits and the classical layers to enhance classification performance. Gradient-based optimization is employed, with quantum circuit gradients computed using the parameter-shift rule as described earlier. This enables efficient, end-to-end training of the hybrid quantum-classical model.

\section{Demonstration}

This section presents experiments demonstrating the feasibility of integrating quantum pointwise convolution with classical models and highlights the superior performance of the quantum model in classification tasks compared to its classical counterpart.

For quantum simulation, we utilized Pennylane to construct quantum circuits and compute their gradients. Quantum circuit optimization was performed on the "default.qubit" simulator using the parameter-shift rule and gradient-based methods. The Adam optimizer was employed with a cross-entropy loss function, a batch size of 128 for training, and 64 for testing. The initial learning rate was set to 0.01, with cosine annealing applied for dynamic adjustment. Training was conducted on NVIDIA Tesla V100 GPUs (16GB) and Intel Xeon CPUs.

\subsection{Classical-Quantum pointwise convolutional model with muti-labels classification on FasionMNIST and CIFAR10}

We utilized a simple convolutional neural network for classification tasks on the FashionMNIST and CIFAR10 datasets as a demo, with 60,000 training images and 10,000 test images for each dataset. 

In the simple classical convolutional network (with three classical convolutional layers expanding to 128 channels), we added the quantum pointwise convolution with 3 strong entanglement layers and classical pointwise convolution to the convolutional neural network respectively, as shown in Fig. 3. We then used fully connected layers for classification and added dropout to prevent overfitting, allowing for a comparison of quantum and classical model performance. This approach helps avoid using excessively deep convolutional networks, which could overshadow the performance improvements brought by pointwise convolution, enabling a more accurate comparison between quantum and classical models.

\begin{figure}[ht]
    \centering
    \includegraphics[width=0.85\columnwidth]{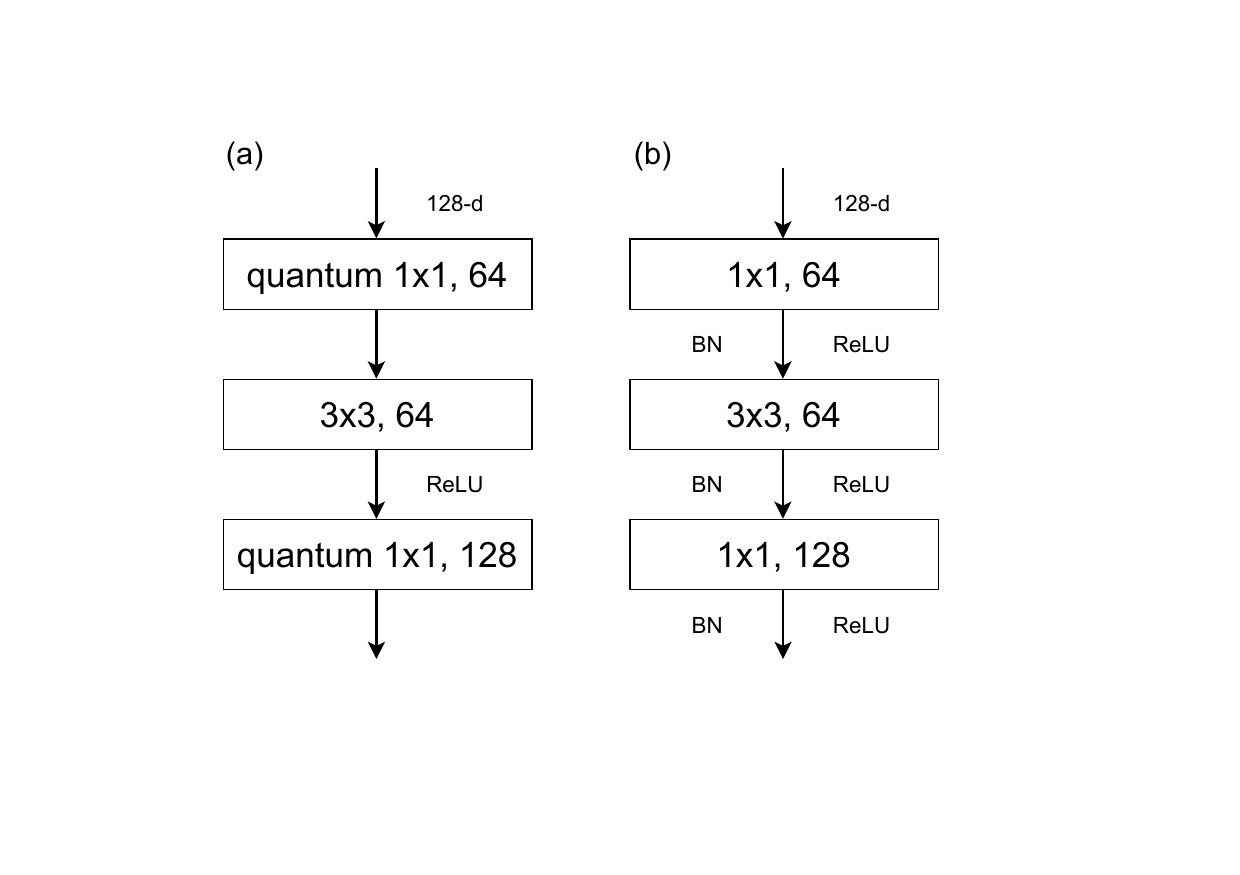}  
    \caption{In (a), the quantum model architecture applies quantum pointwise convolution operations: it starts with a quantum 1x1 convolution layer (64 channels), followed by a classical 3x3 convolution layer (64 channels) with ReLU activation, and concludes with a quantum 1x1 convolution layer that expands to 128 channels. In (b), the classical model begins with a 1x1 convolution layer (64 channels) followed by Batch Normalization (BN) and ReLU. This is followed by a 3x3 convolution layer (64 channels) with BN and ReLU, ending with a 1x1 convolution layer that increases the channels to 128, also followed by BN and ReLU.}
\end{figure}

In the Fig. 4, which shows the traning loss of the epoch, the solid blue line represents the quantum model, while the dashed orange line corresponds to the classical model. Both models exhibit a significant reduction in loss as training progresses, indicating that they are effectively learning and minimizing prediction errors. The quantum model experiences a faster drop in loss in the early epochs and maintains a lower loss throughout the later stages of training, suggesting superior performance in this task compared to the classical model.

In the Fig. 5, which shows the accuracy over the epochs, both models show an upward trend in accuracy as training continues. The quantum model initially outpaces the classical model in accuracy improvement, eventually stabilizing at over 95\% , whereas the classical model stabilizes slightly below this threshold. The quantum model slightly outperforms the classical model in terms of both reducing loss and improving accuracy, especially in the later stages of training. 

\begin{figure}[ht]
    \centering
    \includegraphics[width=0.85\columnwidth]{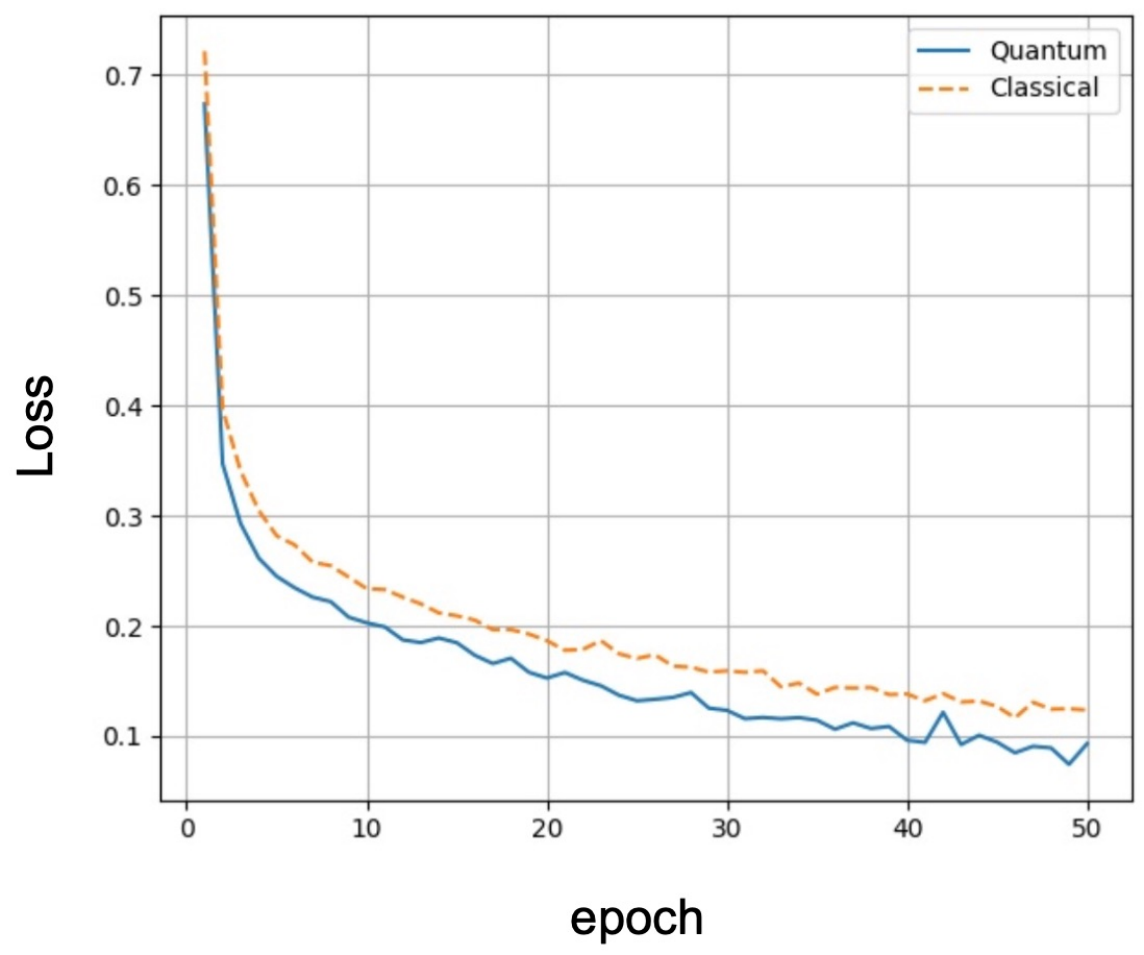}  
    \caption{The figure presents a comparison between the quantum and classical models in terms of loss over the training epochs for classification on the FashionMNIST dataset.}
\end{figure}

\begin{figure}[ht]
    \centering
    \includegraphics[width=0.85\columnwidth]{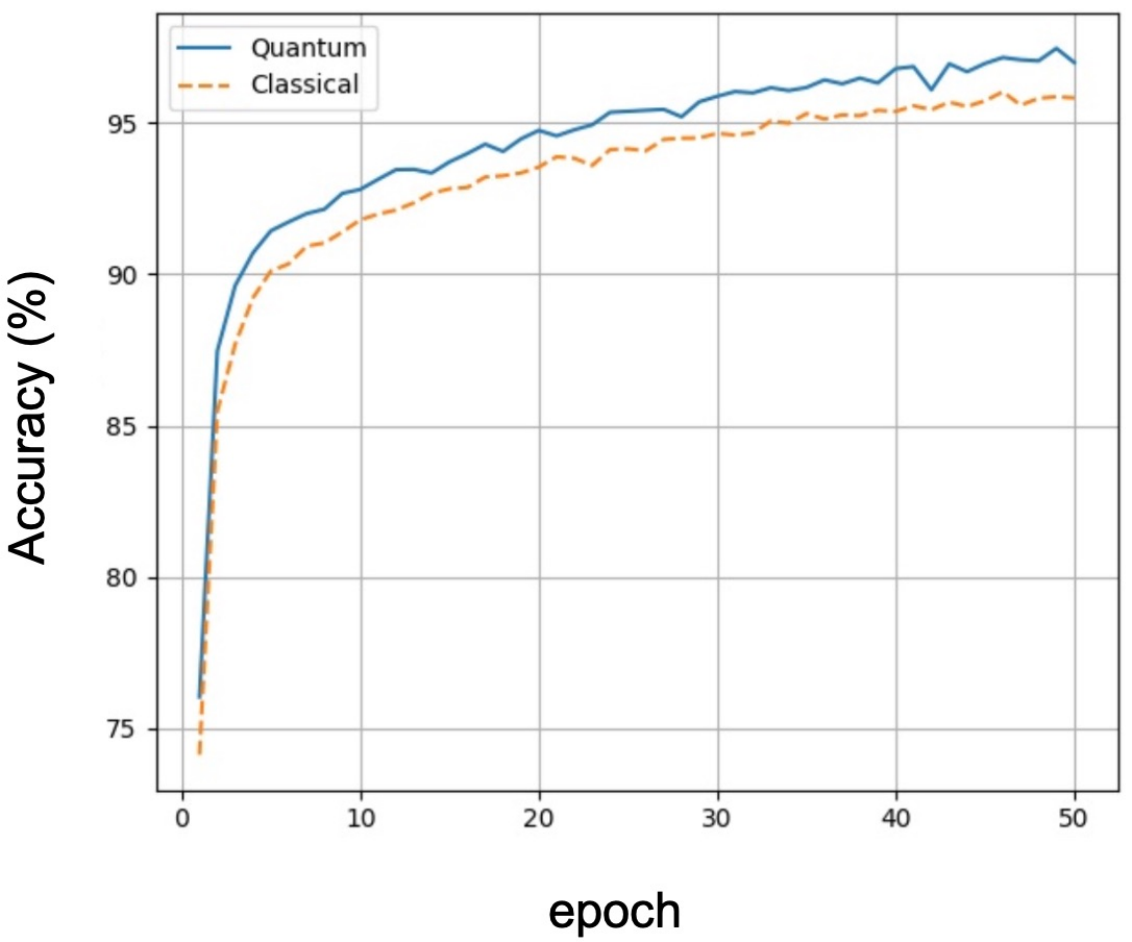}  
    \caption{The figure presents a comparison between the quantum and classical models in terms of accuracy over the training epochs for classification on the FashionMNIST dataset.}
\end{figure}

In the Fig. 6, which shows the traning loss of the epoch, the solid blue line represents the quantum model, and the dashed orange line represents the classical model. Both models experience a significant reduction in loss as training progresses. The quantum model shows a faster decrease in loss and maintains a lower loss throughout the later stages of training, indicating its superiority in the CIFAR10 classification task.

In the Fig. 7, which shows accuracy over the epochs, both models exhibit an increasing trend in accuracy as training continues. The quantum model's accuracy improves more quickly, eventually reaching approximately 90\%, while the classical model stabilizes around 80\%.

\begin{figure}[ht]
    \centering
    \includegraphics[width=0.85\columnwidth]{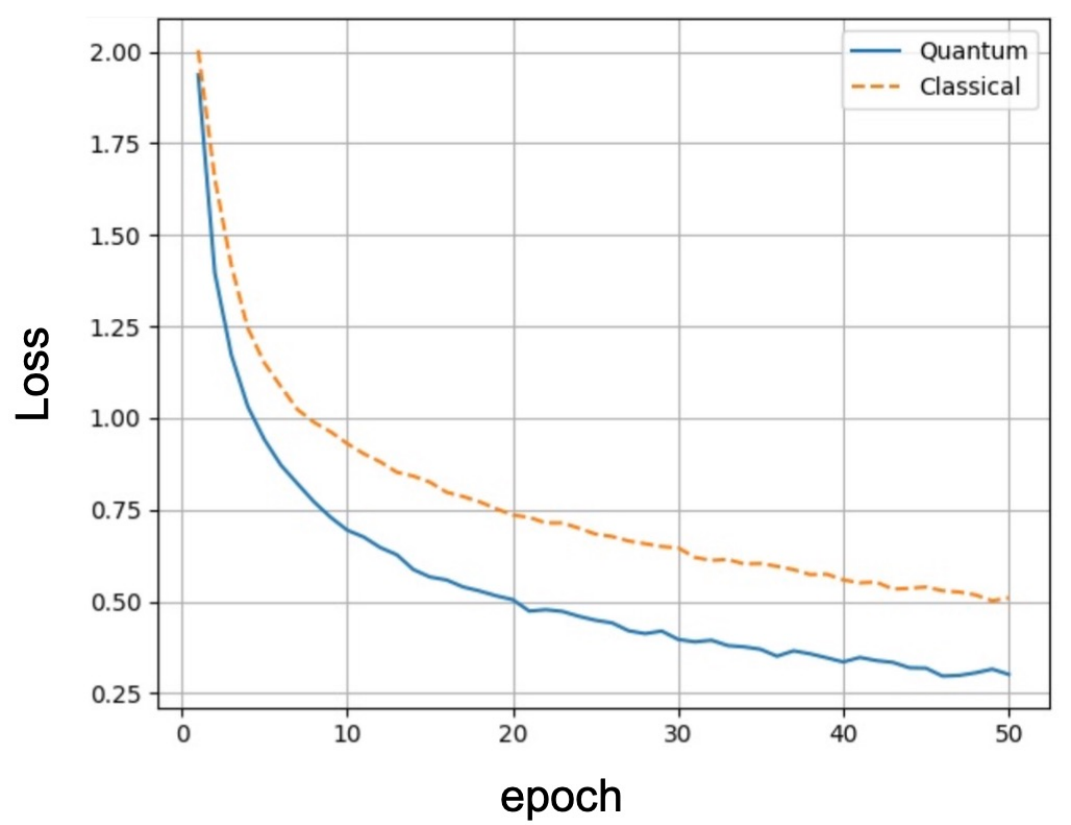}  
    \caption{The figure shows the performance comparison between the quantum and classical models for the CIFAR10 classification task, demonstrating the loss trends over the training epochs.}
\end{figure}

\begin{figure}[ht]
    \centering
    \includegraphics[width=0.85\columnwidth]{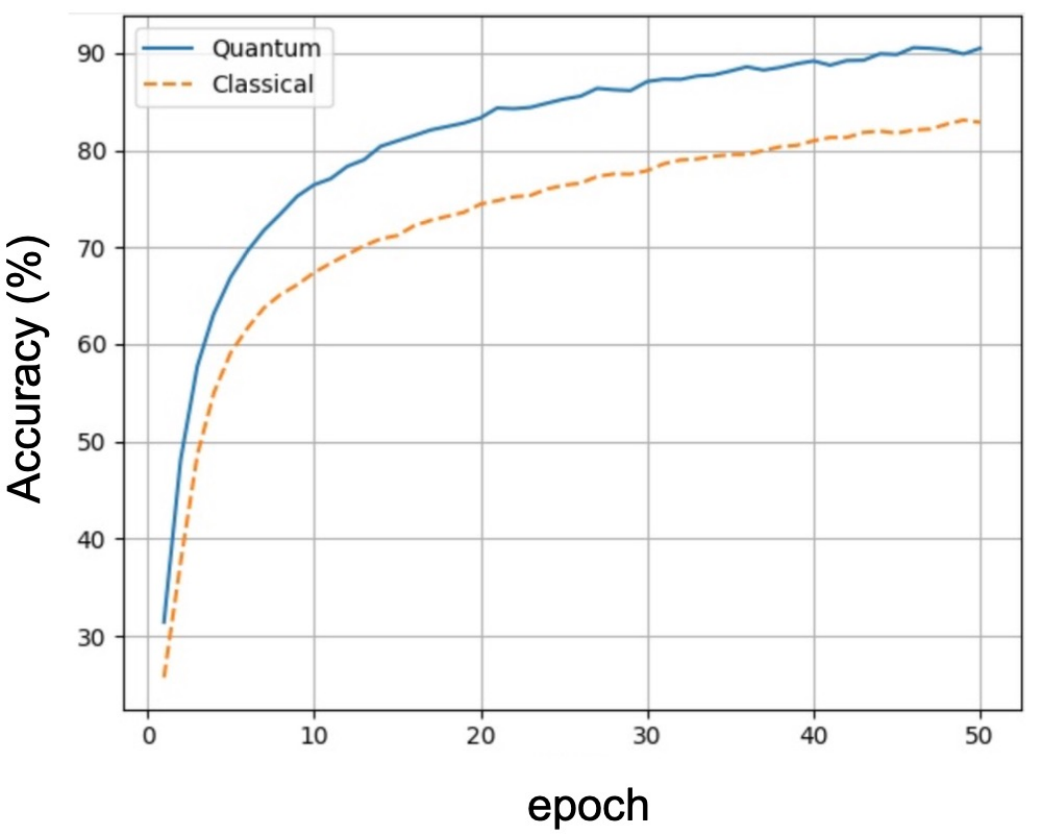}  
    \caption{The figure shows the performance comparison between the quantum and classical models for the CIFAR10 classification task, demonstrating the accuracy trends over the training epochs.}
\end{figure}

\subsection{Comparison between different numbers of strong entangles layers}

In this section, the CIFAR10 dataset was utilized to evaluate the performance of quantum pointwise convolution, same model shown as Fig.3(a), across different layer configurations. This experiment investigates the impact of quantum layer depth on the model’s ability to capture complex features and enhance classification accuracy. By incrementally increasing the number of entanglement layers, we assess whether deeper quantum architectures offer significant performance advantages over shallower ones and how they compare to classical convolutional networks. The results provide critical insights into the scalability of quantum models for real-world tasks such as image classification.

In Fig. 8, quantum models exhibit a steeper accuracy increase during the initial training epochs, surpassing the classical model, which demonstrates slower improvement. Among the quantum configurations, the 4-layer quantum model achieves the highest accuracy, approaching 90\%. The 3-layer and 2-layer quantum models display comparable performance, slightly below the 4-layer model but markedly exceeding the classical model. Although the 1-layer quantum model attains lower accuracy than other quantum configurations, it still outperforms the classical model. The classical model, represented by the dashed purple line, achieves a peak accuracy of approximately 70\%, remaining inferior to all quantum configurations.

In summary, the quantum models, particularly those with 3 and 4 layers, consistently outperform the classical model in both reducing loss and improving accuracy throughout the training process. The deeper quantum models (3 and 4 layers) provide the best performance, suggesting that increasing the depth of quantum convolutional layers contributes to improved classification performance on this task.

\begin{figure}[ht]
    \centering
    \includegraphics[width=0.85\columnwidth]{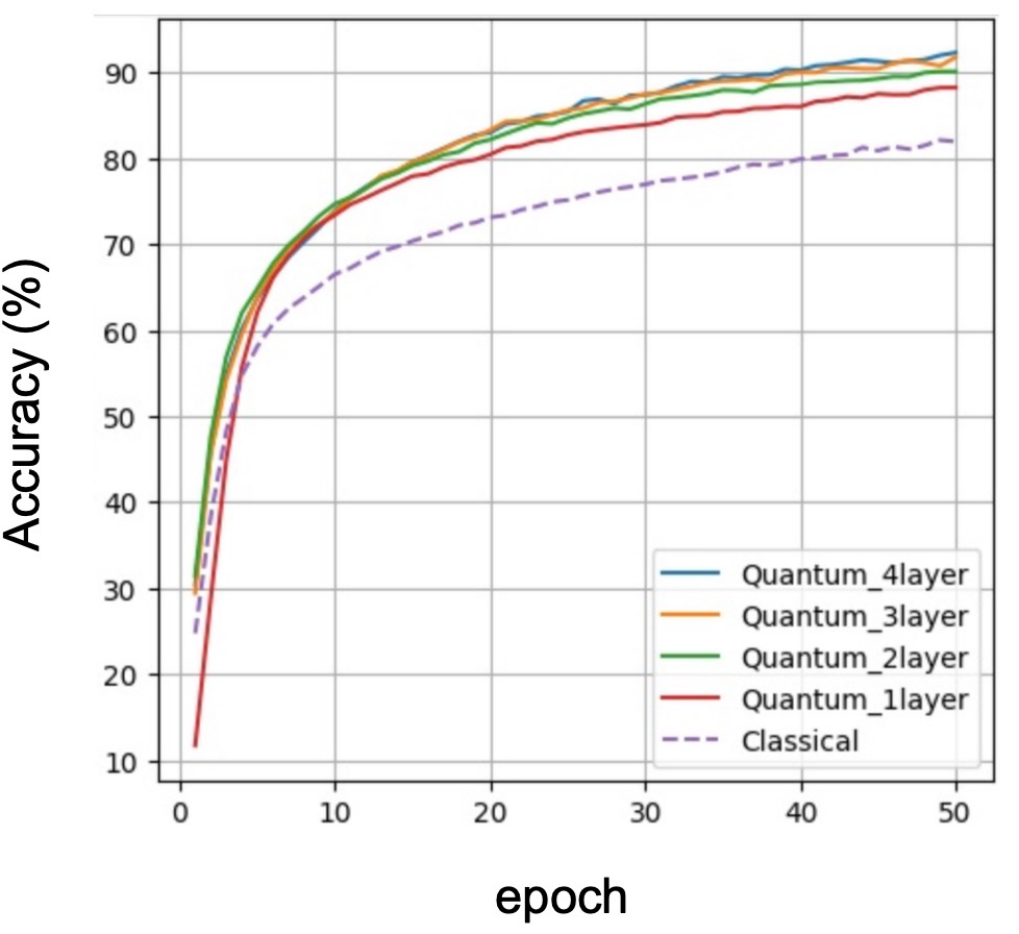}  
    \caption{The figure illustrates the comparison of accuracy over training epochs for various quantum models with different strong entanglement layers, namely 1-layer, 2-layer, 3-layer, and 4-layer quantum models, as well as the classical model. The loss and accuracy are plotted as functions of the epoch, where the solid lines represent the quantum models, and the dashed purple line represents the classical model.}
\end{figure}

\subsection{Discussion}

In this section, we explore potential optimization strategies for the model and examine future directions for applying quantum pointwise convolution, focusing on possible use cases and advancements.

\subsubsection{Optimization}

Each quantum pointwise convolution kernel is represented by a small number of qubits and a relatively simple quantum circuit. Quantum circuit operations are executed on the CPU, while classical convolution operations run on the GPU. However, this approach introduces performance bottlenecks, particularly in execution speed, due to the overhead from CPU-GPU communication. Future work could focus on optimizing the hybrid CPU-GPU architecture to improve execution efficiency.

Additional optimization strategies, such as exploring alternative optimizers, loss functions, and hyperparameter configurations, offer potential for further improvements. In this study, we used the adam optimizer with crossentropy loss to train both the quantum circuit parameters and the classical neural network layers. Investigating different combinations of optimization techniques and hyperparameters could yield insights into enhancing the performance of both quantum and classical components.

\subsubsection{Application}

The improved execution efficiency and compatibility of quantum pointwise convolution with classical models make it a versatile component for integration into various convolutional neural networks (CNNs) or CNN-based architectures. For example, it could replace the pointwise convolution in the Depthwise Separable Convolution layers of MobileNet \cite{b11} or be incorporated into the Bottleneck layers of ResNet \cite{b20}. This flexibility enables quantum pointwise convolution to address more complex tasks and scale to larger models, potentially improving both efficiency and overall performance.

\section{Conclusion}

From quantum circuit design to overall model architecture, we have addressed key bottlenecks and limitations in contemporary quantum neural networks. For data input, we adopt amplitude encoding rather than the commonly used angle embedding \cite{b21}, enabling efficient data embedding with fewer qubits. In the convolutional process, quantum entanglement is employed for both local and global convolution, enhancing the model’s representational capacity by capturing complex feature interactions. Entangled qubits establish correlations across data dimensions, allowing the quantum circuit to process intricate relationships that are challenging for classical networks. In contrast, classical pointwise convolution layers are limited to linear combinations of inputs and lack the ability to model interdependencies across data dimensions, thereby restricting their representational power compared to quantum approaches.

To optimize feature map generation, we leverage quantum circuits’ unique parallelization capabilities. By measuring each qubit individually, a single quantum convolutional kernel generates multiple feature maps, corresponding to the number of qubits, within a single operation. This approach significantly increases efficiency compared to classical convolutional kernels, which typically produce one feature map at a time due to their scalar output nature. This quantum advantage enables compact and efficient feature extraction, reducing model complexity without compromising performance.

In classification experiments, quantum pointwise convolution demonstrated superior representational ability over classical convolutional neural networks of comparable configurations. Substituting classical convolution with quantum circuits yielded equivalent or better performance with significantly fewer parameters. For instance, our quantum model achieved higher accuracy on benchmark tasks while using fewer parameters than the classical model. These findings indicate that quantum-enhanced models provide a promising path to more compact and powerful architectures, suitable for complex tasks and practical applications.

\section{Acknowledgment}

This work was financially supported by the National Science and Technology Council (NSTC), Taiwan, under Grant NSTC 112-2119-M-007-008- and 113-2119-M-007-013-. The authors would like to thank the National Center for High-performance Computing of Taiwan for providing computational and storage resources. The authors also thank Dr. An-Cheng Yang and Dr. Chun-Yu Lin for their support with the hardware environment and for the valuable discussions.

\end{document}